\documentclass[runningheads]{llncs}
\usepackage{booktabs} % For professional looking tables
\usepackage{amssymb}
\usepackage{amsmath}
\usepackage{graphicx}
\usepackage{textcomp}
\usepackage[table]{xcolor} % Add this in your preamble
\usepackage{multirow} 
\usepackage{subcaption}
\usepackage{booktabs} 
\usepackage[misc]{ifsym}
\usepackage{color}
\usepackage{colortbl}
\usepackage{pifont}
\begin{document}
\title{Multimodal Variational Autoencoder for Low-cost Cardiac Hemodynamics Instability Detection}

\newcommand{\corrauth}{\textsuperscript{(\Letter)}}
\titlerunning{Multimodal Variational Autoencoder for Cardiac Hemodynamics}

\author{Mohammod N. I. Suvon\inst{1,2}\corrauth \and Prasun C. Tripathi\inst{1,7} \and Wenrui Fan\inst{1,2} \and Shuo Zhou\inst{1,2} \and Xianyuan Liu\inst{1,2} \and Samer Alabed\inst{3,4,5} \and Venet Osmani\inst{6} \and Andrew J. Swift\inst{3,4,5} \and Chen Chen\inst{1,8,9} \and Haiping Lu\inst{1,2,5}} 
% index{Suvon, Mohammod}, index{Tripathi, Prasun}, index{Fan, Wenrui}, index{Zhou, Shuo}, index{Liu, Xianyuan}, index{Alabed, Samer}, index{Osmani, Venet}, index{Swift, Andrew}, index{Chen, Chen}, index{Lu, Haiping}

\authorrunning{Suvon et al.}

\institute{Department of Computer Science, University of Sheffield, Sheffield, UK \and Centre for Machine Intelligence, University of Sheffield, Sheffield, UK \and Department of Infection, Immunity and Cardiovascular Disease, University of Sheffield, Sheffield, UK \and  Department of Clinical Radiology, Sheffield Teaching Hospitals, Sheffield, UK \and INSIGNEO, Institute for in Silico Medicine, University of Sheffield, Sheffield, UK \and 
Information School, University of Sheffield, Sheffield, UK \and 
Department of Computer Science, IITRAM, Gujarat, India\and Department of Engineering Science, University of Oxford, Oxford, UK \and  Department of Computing, Imperial College London, London, UK \\
\email{\{m.suvon\corrauth, p.c.tripathi , wenrui.fan, shuo.zhou, xianyuan.liu, s.alabed, v.osmani, a.j.swift, chen.chen2, h.lu\}@sheffield.ac.uk}}

%\author{Prasun C. Tripathi\inst{1}\corrauth \and Mohammod N. I. Suvon\inst{1} \and Lawrence Schobs\inst{1} \and Shuo Zhou\inst{1,2} \and Samer Alabed\inst{3,4,5} \and Andrew J. Swift\inst{3,4,5} \and Haiping Lu\inst{1,2,5}} 

%\author{MNI Suvon}\inst{1,2}
%\author{Lawrence Schob}\inst{3}

%\authorrunning{Tripathi et al.}
%If there are more than two authors, 'et al.' is used.

%\institute{Department of Computer Science, University of Sheffield, Sheffield, UK \and Centre for Machine Intelligence, University of Sheffield, Sheffield, UK \and Department of Infection, Immunity and Cardiovascular Disease, University of Sheffield, Sheffield, UK \and  Department of Clinical Radiology, Sheffield Teaching Hospitals, Sheffield, UK \and INSIGNEO, Institute for in Silico Medicine, University of Sheffield, Sheffield, UK \\
%\email{\{p.c.tripathi\corrauth , m.suvon, laschobs1, shuo.zhou, s.alabed, a.j.swift, h.lu\}@sheffield.ac.uk}}
%\and
%Second Author\inst{2,3}\orcidID{1111-2222-3333-4444} \and
%Third Author\inst{3}\orcidID{2222--3333-4444-5555}}
%\institute{Anonymous \and
%Springer Heidelberg, Tiergartenstr. 17, 69121 Heidelberg, Germany
\maketitle              % typeset the header of the contribution
\begin{abstract}

%Pulmonary Arterial Wedge Pressure (PAWP), a key marker of cardiac hemodynamic instability, is typically assessed via Right Heart catheterization (RHC). However, non-invasive methods are preferred clinically for screening high-risk individuals at a population level. The research efforts in this direction are mainly devoted to single modalities (e.g., cardiac MRI, Echocardiography) with limited number of patients data. In this work, we develop a Multimodal Variational Auto-encoder (mVAE) that leverages two heterogeneous modalities Electrocardiogram (ECG) and cardiac X-ray (CXR). ECG modality provides temporal features for the prediction, whereas CXR modality offers anatomical features. We pre-train our mVAE using an unlabelled MIMIC dataset of 50982 subjects to learn domain specific prior knowledge and then fine-tune on the labelled dataset of the ASPIRE registry of 795 subjects who underwent the RHC procedure. We not only incorporate shared features in the latent space but also add unique modality-specific features by a novel Evidence Lower Bound (ELBO) loss. Additionally, we design a Multimodal Linear Autoencoder (mLAE) which offers more interpretability benefits for clinical decision-making. Extensive evaluations with respect to existing methods show that our linear and non-linear auto-encoders offer promising performance.

Recent advancements in non-invasive detection of cardiac hemodynamic instability (CHDI) primarily focus on applying machine learning techniques to a single data modality, e.g. cardiac magnetic resonance imaging (MRI). Despite their potential, these approaches often fall short especially when the size of labeled patient data is limited, a common challenge in the medical domain. Furthermore, only a few studies have explored multimodal methods to study CHDI, which mostly rely on costly modalities such as cardiac MRI and echocardiogram. In response to these limitations, we propose a novel multimodal variational autoencoder ($\text{CardioVAE}_\text{X,G}$) to integrate low-cost chest X-ray (C\textbf{X}R) and electrocardiogram (EC\textbf{G}) modalities with pre-training on a large unlabeled dataset. Specifically, $\text{CardioVAE}_\text{X,G}$ introduces a novel tri-stream pre-training strategy to learn both shared and modality-specific features, thus enabling fine-tuning with both unimodal and multimodal datasets. We pre-train $\text{CardioVAE}_\text{X,G}$ on a large, unlabeled dataset of $50,982$ subjects from a subset of MIMIC database and then fine-tune the pre-trained model on a labeled dataset of $795$ subjects from the ASPIRE registry. Comprehensive evaluations against existing methods show that $\text{CardioVAE}_\text{X,G}$ offers promising performance  (AUROC $=0.79$ and Accuracy $=0.77$), representing a significant step forward in non-invasive prediction of CHDI. Our model also excels in producing fine interpretations of predictions directly associated with clinical features, thereby supporting clinical decision-making.

%Our model is also capable of producing fine interpretations of predictions linked to clinical features for clinical decision-making. 

%for clinical decision-making.

\end{abstract}
\section{Introduction}

Cardiac hemodynamic instability (CHDI) can lead to unreliable and inefficient cardiovascular function and even heart failure. Pulmonary Artery Wedge Pressure (PAWP) is an important surrogate marker for detecting the severity of CHDI and heart failure. Elevated PAWP indicates left ventricular filling pressure and reduced contractility of the heart~\cite{garg2022cardiac}. It can be precisely measured by invasive and expensive right heart catheterization (RHC). However, simpler and non-invasive methods are often required to monitor critical patients. In recent years, several machine learning-based and/or deep learning-based methods were developed for PAWP prediction from medical images acquired by non-invasive technique~\cite{garg2022cardiac,traversi2001doppler,tripathi2023tensor}. It has been shown that it is possible to predict PAWP not only from high-cost, high precision scans such as cardiac magnetic resonance imaging (MRI)~\cite{garg2022cardiac} and echocardiography~\cite{traversi2001doppler}, but also from more affordable, accessible scans or measurements such as chest X-rays (CXR)~\cite{kusunose2020deep,kusunose2023deep,hirata2021deep} and electrocardiogram (ECG)~\cite{raghu2023ecg,schlesinger2022deep}. 

% Most aforementioned works focus mainly on extracting measurements from \emph{one, single} data modality. More recently, multi-modal learning have demonstrated superior diagnostic performance compared to those uni-modality learning approaches~\cite{vafaii2024multimodal,zhou2023transformer,suvon2022multimodal}. One representative work is from~\cite{tripathi2023tensor}, where a tensor-based multimodal pipeline is developed to combine features from expensive cardiac MRI images and the measurements from [XXX modality] for higher precision. In this work, different from prior works, we use only low-cost modalities, e.g., CXR and ECG and develop a multi-modal method to improve the PAWP prediction performance. We believe that such a low-cost approach is a more promising and feasible for broader adoption in low-income countries where financial constraints limit the acquisition of expensive equipment, such as MRI scanners. Yet, there are two main challenges remained: a) Compared to MRI and ultrasound, the information from 2D CXR is relatively limited; b) the labelled dataset is often limited, especially for our task. This is because obtaining precise PAWP requires \emph{invasive} approach to detect the true PWAP value, which is not always possible.

Most aforementioned studies focus mainly on extracting measurements from \emph{one, single} data modality. More recently, multimodal learning has demonstrated superior diagnostic performance to unimodal learning approaches~\cite{vafaii2024multimodal,zhou2023transformer,suvon2022multimodal}. For example, Tripathi et al.~\cite{tripathi2023tensor} developed a tensor-based multimodal pipeline to combine features from cardiac MRI imaging with cardiac measurement for higher precision. In this work, we developed a multimodal model utilizing only low-cost modalities, e.g., CXR and ECG, to improve the performance of PAWP prediction. Such a low-cost approach is more relevant and feasible for broader adoption in low-income countries with limited access to MRI scans. The two main challenges in this low-cost approach are a) the limited information in 2D CXR images and 1D ECG signals compared to MRI and ultrasound and b) the scarcity of PAWP measurements in the labeled dataset relating to the difficulty and expense of obtaining precise and \emph{invasive} PAWP.

\begin{figure}[!t]
  \centering
  \includegraphics[width=\textwidth, trim=8cm 5cm 8cm 2.5cm]{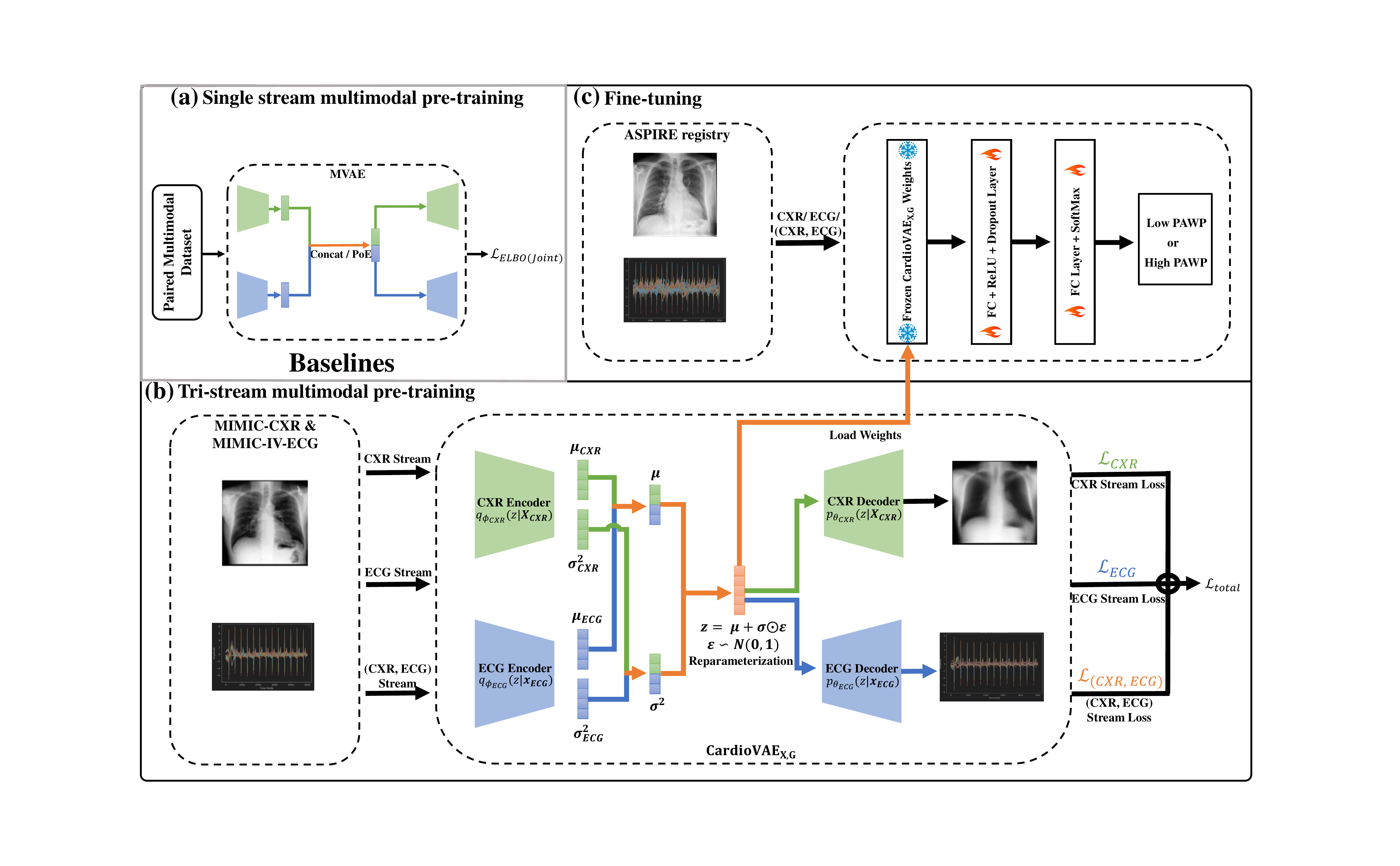}
  \caption{The baselines and proposed $\text{CardioVAE}_\text{X,G}$ for PAWP prediction. (a) Top left: the single stream MVAE baselines~\cite{wu2018multimodal,li2023towards} require pre-training on paired data (CXR, ECG) and utilize only the shared features using concatenation or product of expert (PoE)~\cite{hinton2002training} based multimodal fusion methods. (b) Bottom: Tri-stream multimodal pre-training that can learn both shared and modality-specific features. (c) Top right: fine-tuning on ASPIRE registry. Due to the tri-stream flow, our model can be fine-tuned on a single modality or both modalities.}
  \label{fig1}
\end{figure}

To tackle the above challenges, we first develop a cardiac multimodal variational autoencoder ($\text{CardioVAE}_\text{X,G}$), aiming at maximizing the value from C\textbf{X}R images and EC\textbf{G}s through joint training. Specifically, $\text{CardioVAE}_\text{X,G}$ is first pre-trained using a novel tri-stream multimodal pre-training strategy, leveraging the value from a large-scale public, \emph{unlabeled} paired CXR and ECG data to reduce the need for large-scale, \emph{labeled} data. We then fine-tune the pre-trained model on a relatively small dataset for the PAWP prediction task. The \textbf{main contributions} of this paper are three-fold:
1) We introduce $\text{CardioVAE}_\text{X,G}$, a novel multimodal variational autoencoder (Fig.~\ref{fig1}) for low-cost, non-invasive PAWP prediction. Unlike previous approaches that rely on expensive modalities such as cardiac MRI, $\text{CardioVAE}_\text{X,G}$ efficiently integrates lower-cost CXR and ECG modalities. 
2) We develop a novel tri-stream pre-training strategy with $\text{CardioVAE}_\text{X,G}$ model to learn both shared and modality-specific features. This approach enables our $\text{CardioVAE}_\text{X,G}$ model to be fine-tuned with unimodal or multimodal datasets, greatly improving usability. 
3) We performed extensive experiments to show the promising performance of our model. Another unique feature is that $\text{CardioVAE}_\text{X,G}$ is capable of providing explainable feature visualization for the interpretation of the clinical decisions (see Fig.~\ref{fig:both_interpret}), enhancing their applicability and reliability in real-world scenarios.
%This feature can facilitate the use of datasets even when one modality may be absent (which is common in medical domain), broadening the potential for application.
%3) We pre-trained $\text{CardioVAE}_\text{X,G}$ on an extensive, unlabeled dataset from a subset of MIMIC database~\cite{johnson2019mimic,gowmimic}, encompassing $50,982$ subjects, and subsequently fine-tuned it on an external labelled dataset of $795$ subjects from the ASPIRE registry~\cite{hurdman2012aspire}. We performed extensive experiments and comparisons with existing methods to demonstrate the promising performance of our model (AUC=$0.790 \pm 0.03$ and Accuracy=$0.772 \pm 0.04$). Another unique feature is that $\text{CardioVAE}_\text{X,G}$ is capable of providing explainable feature visualization for the interpretation of the clinical decisions (see Fig.~\ref{fig:both_interpret}), enhancing their applicability and reliability in real-world scenarios.

\section{Methods}
\label{s2}
%\textcolor{red}{
The proposed $\text{CardioVAE}_\text{X,G}$ for PAWP prediction is depicted in Fig.~\ref{fig1}. It leverages low-cost CXR and ECG modalities for the prediction. Our model utilizes tri-stream multimodal pre-training, incorporating data streams from three sources: CXR, ECG, and the modality pair (CXR, ECG). Tri-stream setting allows our model to learn modality-specific features along with shared features in contrast with the baseline MVAE~\cite{wu2018multimodal} that utilizes only the shared features of CXR and ECG. In the following, we discuss each building block of our model.

%n the fine-tuning, we freeze $\text{CardioVAE}_\text{X,G}$ weights and only train two Fully Connected (FC) layers.  
\noindent\textbf{CXR Encoder:} Our CXR encoder incorporates a convolutional neural network (CNN) consisting of three Convolutional (CONV) layers. It processes a CXR image \(\mathbf{X}_\text{CXR} \in \mathbb{R}^{H \times W \times C}\), where \(H\), \(W\), and \(C\) represent the height, width, and number of channels, respectively. Each CONV layer is configured with a \(3 \times 3\) kernel, a stride of $2$, and padding of $1$. We use channel depths of $16$, $32$, and $64$ in CONV layers. Following this, the data is flattened and passed through two Fully Connected (FC) layers to produce the mean ($\boldsymbol{\mu}_\mathbf{{\text{CXR}}}$) and variance ($\boldsymbol{\sigma^2}_\mathbf{{\text{CXR}}}$) vectors, which are crucial for defining the CXR image's latent space representation and establishing its approximate posterior distribution parameters $q_{\phi_\text{CXR}}(z|\mathbf{X}_{\text{CXR}})$.

\noindent\textbf{ECG Encoder:} The ECG encoder comprises three one-dimensional (1D) CONV layers. It processes the sequential ECG signal \(\mathbf{x}_\text{ECG} \in \mathbb{R}^{L}\), where \(L\) denotes the length of the signal. Each 1D CONV layer is configured with a kernel size of $1 \times 3$, a stride of $2$, and padding of $1$. We use channel depths of $16$, $32$, and $64$ in three CONV layers. Following this, the signal is flattened and processed through two FC layers, yielding the mean ($\boldsymbol{\mu}_\mathbf{{\text{ECG}}}$) and variance ($\boldsymbol{\sigma^2}_\mathbf{{\text{ECG}}}$) which are crucial for the ECG signal's latent space representation and its approximate posterior distribution $q_{\phi_\text{ECG}}(z|\mathbf{x}_{\text{ECG}})$.

\noindent\textbf{Multimodal Integration:} In multimodal integration, we employ a Product of Experts (PoE)~\cite{hinton2002training} approach to combine the approximate posterior distributions from the CXR and ECG encoders with a standard Gaussian prior \(p(z) = \mathcal{N}(0, \mathbf{I})\) into a unified latent space, effectively synthesizing the individual expert opinions, where \(\mathbf{I}\) is the identity matrix. The combined mean \(\boldsymbol{\mu}\) and variance \(\boldsymbol{\sigma^2}\) of the latent space~\cite{hwang2021multi} are computed as: $\boldsymbol{\mu} = (\sum_{m = 1}^M \boldsymbol{\mu_m}/\boldsymbol{\sigma_m^2})/(\sum_{m = 1}^M 1/\boldsymbol{\sigma_m^2})$ and $\boldsymbol{\sigma^2} = 1/(\sum_{m = 1}^M1 / \boldsymbol{\sigma_m^2})$, where \(m\) represents the modality, ranging from 1 to \(M\). This leads to the calculation of the latent space variable $\mathbf{z}$ using the reparameterization trick $\mathbf{z} = \boldsymbol{\mu} + \boldsymbol{\sigma} \cdot \boldsymbol{\epsilon}$, where $\boldsymbol{\epsilon}$ is drawn from $\mathcal{N}(0, \mathbf{I})$. This step allows the decoders to generate diverse yet consistent reconstructions, effectively establishing the approximate joint posterior distribution $q_{\phi_\text{CXR,ECG}}(z|\mathbf{X_{\text{CXR}}},\mathbf{x_{\text{ECG}}})$ and reflecting a deep understanding of the merged input data.

%++add that the PoE desing in a way where if the modalitis are one then it will only prodcut with the gaussian prior and go though the repramatization trick to latent space.
\noindent\textbf{Decoder Design:} The CXR and ECG decoders aim to reconstruct inputs from the latent variable \(\mathbf{z}\), sampled from the unified latent space learned by the PoE~\cite{hinton2002training} approach. For CXR image reconstruction, the CXR decoder $p_{\theta_\text{CXR}}(\mathbf{X_{\text{CXR}}}|z)$  is a CNN consisting of three transposed convolutional layers, effectively restoring sample $\mathbf{X_{\text{CXR}}}$ from latent variable $z$. Similarly, the ECG decoder $p_{\theta_\text{ECG}}(\mathbf{x_{\text{ECG}}}|z)$ utilizes three 1D transposed convolution layer to convert \(\mathbf{z}\) into precise temporal waveform $\mathbf{x_{\text{ECG}}}$. This reconstruction process leverages the inherent flexibility of the Gaussian distribution within the latent space, enabling the production of diverse, high-fidelity reconstructions by sampling various \(\mathbf{z}\) points.

%The $\text{CardioVAE}_\text{X,G}$ is trained by maximizing the Evidence Lower Bound (ELBO)~\cite{kingma2013auto} defined as:
%\begin{equation}
%\mathcal{L}_\text{ELBO} = \mathbb{E}_{q_{\phi}(z|x)}[ \log p_\theta(x|z)] -  D_\text{KL}[q_{\phi}(z|x) || p(z)],
%\end{equation}
%where $\mathbb{E}_{q_{\phi}(z|x)}[\log p_\theta(x|z)]$ is the expected conditional log-%likelihood of the data given the latent variable z, indicating how well the model reconstructs %the data. $D_\text{KL}[q_{\phi}(z|x) || p(z)]$ is the Kullback-Leibler (KL) %divergence~\cite{bu2018estimation} between the approximate posterior $q_{\phi}(z|x)$ and the %prior distribution over the latent variables $p(z)$, serving as a regularization term. The %ELBO loss aims to maximize the likelihood of the data while minimizing the difference between %the approximate posterior and the prior. 

\noindent\textbf{Tri-stream Pre-training Strategy:} For pre-training, we  extend~\cite{wu2018multimodal,joy2021learning} to introduce a tri-stream strategy that processes CXR and ECG data both separately and jointly to capture the unique and shared features of each modality in the latent space. We pass three data streams through our $\text{CardioVAE}_{\text{X,G}}$ model: 1) CXR only, 2) ECG only, and 3) paired CXR and ECG, and calculate three separate losses for each stream by incorporating Evidence Lower Bound (ELBO)~\cite{kingma2013auto}. We define $\mathcal{L}_\text{CXR}$ for CXR, $\mathcal{L}_\text{ECG}$ for ECG, and $\mathcal{L}_\text{(CXR,ECG)}$ for joint loss, as follows:
\begin{equation}
    \mathcal{L}_\text{CXR} = \mathbb{E}_{q_{\phi_\text{CXR}}(z|\mathbf{X}_{\text{CXR}})}[\lambda_{\text{CXR}} \log p_{\theta_\text{CXR}}(\mathbf{X}_{\text{CXR}}|z)] - \beta D_{\text{KL}}[q_{\phi_\text{CXR}}(z|\mathbf{X}_{\text{CXR}}) || p(z)],
\label{elbo_CXR}
\end{equation}
\begin{equation}
    \mathcal{L}_\text{ECG} =\mathbb{E}_{q_{\phi_\text{ECG}}(z|\mathbf{x}_{\text{ECG}})}[\lambda_{\text{ECG}} \log p_{\theta_\text{ECG}}(\mathbf{x}_{\text{ECG}}|z)] - \beta D_{\text{KL}}[q_{\phi_\text{ECG}}(z|\mathbf{x}_{\text{ECG}}) || p(z)],
\label{elbo_ECG}
\end{equation}
\begin{align}
    \mathcal{L}_\text{(CXR,ECG)} = \mathbb{E}_{q_{\phi_\text{CXR}}(z|\mathbf{X}_{\text{CXR}})}[\lambda_{\text{CXR}} \log p_{\theta_\text{CXR}}(\mathbf{X}_{\text{CXR}}|z)] \nonumber\\
+ \mathbb{E}_{q_{\phi_\text{ECG}}(z|\mathbf{x}_{\text{ECG}})}[\lambda_{\text{ECG}} \log p_{\theta_\text{ECG}}(\mathbf{x}_{\text{ECG}}|z)] \nonumber\\
- \beta D_{\text{KL}}[q_{\phi_\text{(CXR,ECG)}}(z|\mathbf{X_{\text{CXR}}},\mathbf{x_{\text{ECG}}}) || p(z)],
\label{elbo_CXR_ECG}
\end{align}
% \begin{multline}
% \scriptsize
% \mathcal{L}_\text{CXR} = \mathbb{E}_{q_{\phi_\text{CXR}}(z|\mathbf{X}_{\text{CXR}})}[\lambda_{\text{CXR}} \log p_{\theta_\text{CXR}}(\mathbf{X}_{\text{CXR}}|z)] \\- \beta D_{\text{KL}}[q_{\phi_\text{CXR}}(z|\mathbf{X}_{\text{CXR}}) || p(z)],
% \label{elbo_CXR}
% \end{multline}
% \begin{multline}
% \scriptsize
% \mathcal{L}_\text{ECG} =\mathbb{E}_{q_{\phi_\text{ECG}}(z|\mathbf{x}_{\text{ECG}})}[\lambda_{\text{ECG}} \log p_{\theta_\text{ECG}}(\mathbf{x}_{\text{ECG}}|z)] \\- \beta D_{\text{KL}}[q_{\phi_\text{ECG}}(z|\mathbf{x}_{\text{ECG}}) || p(z)],
% \label{elbo_ECG}
% \end{multline}
% \begin{multline}
% \scriptsize
% \mathcal{L}_\text{(CXR,ECG)} = \mathbb{E}_{q_{\phi_\text{CXR}}(z|\mathbf{X}_{\text{CXR}})}[\lambda_{\text{CXR}} \log p_{\theta_\text{CXR}}(\mathbf{X}_{\text{CXR}}|z)] \\
% + \mathbb{E}_{q_{\phi_\text{ECG}}(z|\mathbf{x}_{\text{ECG}})}[\lambda_{\text{ECG}} \log p_{\theta_\text{ECG}}(\mathbf{x}_{\text{ECG}}|z)] \\
% - \beta D_{\text{KL}}[q_{\phi_\text{(CXR,ECG)}}(z|\mathbf{X_{\text{CXR}}},\mathbf{x_{\text{ECG}}}) || p(z)],
% \label{elbo_CXR_ECG}
% \end{multline}
where the first part $\mathbb{E}_{q_{\phi}}$ of the losses is the expected conditional log-likelihood of the data given the latent variable \(z\), indicating how well the model reconstructs the data. This part takes the reconstructed output from the decoder and finds the reconstruction error. The second part of the losses, \(D_{\text{KL}}\), is the Kullback-Leibler (KL) divergence~\cite{bu2018estimation} between the approximate posterior (CXR or ECG or CXR-ECG) and the prior distribution \(p(z)\), following the Gaussian distribution, over the latent variables, calculated in encoders and multimodal integration phase. These losses aim to maximize the likelihood of the data while minimizing the difference between the approximate posterior and the prior. Moreover, in our losses, we incorporate two important hyperparameters \(\lambda\)~\cite{lawry2023multi} to balance the weights of modalities and \(\beta\)~\cite{higgins2016beta} to balance the trade-off between reconstruction loss and KL divergence. In practice, \(\lambda = 1\) and \(\beta\) are slowly annealed from \(0\) to \(1\) to form a valid lower bound on the evidence~\cite{alemi2018fixing}. After calculating three losses in Eq.~(\ref{elbo_CXR}-\ref{elbo_CXR_ECG}) for the three streams, we combine them to obtain a total loss $\mathcal{L}_\text{total}$ as follows: 
\begin{equation}
\mathcal{L}_\text{total} = \mathcal{L}_{\text{CXR}} + \mathcal{L}_{\text{ECG}} + \mathcal{L}_{\text{(CXR,ECG)}}.
\end{equation}

The tri-stream methodology enhances the model's capability to accurately represent and reconstruct multimodal data, leveraging the strengths of each modality towards a comprehensive representation of the multimodal inputs for improved overall performance.

\noindent\textbf{Fine-tuning Strategy:} Applying the pre-trained model to the CXR-ECG classification task is straightforward, as tri-stream pre-training enables the model to learn both modality-specific and shared features. During fine-tuning, the new downstream dataset is passed through the frozen model to extract features which are subsequently processed by two FC layers. The binary cross-entropy loss~\cite{goodfellow2016deep} is used for fine-tuning our model.

%Applying the pre-trained model to the CXR-ECG classification task is straightforward, as tri-stream pre-training enables the model to learn both modality-specific and shared features. During fine-tuning, the pre-trained model is kept frozen to preserve the earlier features it has learned. The new downstream dataset is processed through the frozen model to extract features. These features are then processed by two FC layers: the first FC layer adapts the extracted features from the frozen model for classification, and the second FC layer finalizes the class prediction. The binary cross-entropy loss~\cite{goodfellow2016deep} has been utilized during the fine-tuning.

\section{Experimental Results and Analysis}
\noindent\textbf{Dataset for Pre-training}: 
We pre-trained our $\text{CardioVAE}_\text{X,G}$ model using two datasets, MIMIC-CXR~\cite{johnson2019mimic} and MIMIC-IV-ECG~\cite{gowmimic}, by pairing them via unique patient ID and time. This gave us $50,982$ pairs of CXR-ECG samples. 

\noindent\textbf{Study Population and Dataset for Downstream Task}: 
We evaluated all models using a dataset from the ASPIRE registry~\cite{hurdman2012aspire} for the detection of CHDI via PAWP prediction in patients with suspected pulmonary hypertension. The local institutional review board and ethics committee approved this study. A total of $795$ patients who underwent RHC, CXR, and ECG were included in this study. Based on the measurements from RHC (using a balloon-tipped $7.5$ French thermodilution catheter), we found $560$ patients with normal PAWP (\(\leq 15\) mmHg), and  $235$ with elevated PAWP (\(> 15\) mmHg). Table~\ref{data} summarizes the patient characteristics of the used ASPIRE registry dataset. 

\begin{table}[!t]
\caption{Patients characteristics of included patients in ASPIRE registry dataset. $p$-values were obtained using $t$-test~\cite{welch1947generalization}.}\label{data}
\centering
\scalebox{0.90}{
\begin{tabular}{l|c|c|c}
\hline
& Low PAWP($\le15$) & High PAWP($>15$) & $p$-value\\
\hline
Number of patients&$560$&$235$& -\\
\hline
Age (in years)&$58.98\pm 15.30$&$62.62\pm 13.75$& $0.0017$\\
\hline
Body Surface Area (BSA) &$1.92\pm 0.26$&$1.98\pm 0.24$& $0.0025$\\
\hline
Heart Rate (bpm) &$78.87\pm 13.87$&$74.75\pm 14.93$& $0.0002$\\
\hline
Pulmonary Artery Pressure &$42.78\pm 14.01$&$44.86\pm 12.31$& $0.0483$\\
\hline
Pulmonary Artery Systolic Pressure &$71.35\pm 24.00$&$75.86\pm 23.75$& $0.0155$ \\
\hline
Pulmonary Artery Diastolic Pressure &$25.06\pm 10.13$&$26.82\pm 7.87$& $0.0176$ \\
\hline
PAWP (mmHg) &$9.94\pm 3.06$&$19.42\pm 3.51$& $<0.0001$\\
\hline
\end{tabular}}
\end{table}

% \textcolor{red}{(Samer and Andy (Procedure of obtaining the CXR and ECG)}

% \begin{figure}[t]
% \centering
% \begin{subfigure}{.46\textwidth}
%   \centering
%   \includegraphics[width=.47\linewidth]{orginal_CXR.png}
%   \includegraphics[width=.47\linewidth]{orginal_ECG.png}
%   \caption{Original}
%   \label{fig:original}
% \end{subfigure}
% \hfill % Use \hfill to add space only if needed
% \begin{subfigure}{.46\textwidth}
%   \centering
%   \includegraphics[width=.47\linewidth]{recon_CXR.png}
%   \includegraphics[width=.47\linewidth]{recon_ECG.png}
%   \caption{Reconstructed}
%   \label{fig:reconstructed}
% \end{subfigure}
% \caption{Qualitative Comparison}
% \label{fig:test}
% \end{figure}

\noindent\textbf{Experimental Design:}
We converted CXR and 12-lead ECG data to $2$D images ($224\times 224$) and $1$D signals ($1\times 60,000$), respectively. We pre-trained our model with unlabeled MIMIC subset~\cite{johnson2019mimic,gowmimic} by partitioning it with a $90:10$ ratio for training and validation sets. The hyperparameters for \(\lambda_\text{CXR}\) and \(\lambda_\text{ECG}\) were selected using grid search. The optimal hyperparameters were then used to pre-train $\text{CardioVAE}_\text{X,G}$ model on the whole MIMIC subset. We used the Fréchet inception distance~\cite{obukhov2020quality} to assess the performance of pre-training. For pre-training, we used Adam optimizer with a learning rate of $0.001$ and a batch size of $128$, and trained for $100$ epochs to ensure the model convergence.

For fine-tuning, we froze the layers in the encoders and fine-tuned FC layers on the ASPIRE registry dataset. We used $128$ nodes in FC layer and used a dropout of $0.5$. We evaluated the prediction performance using $10$-fold cross-validation with a training and validation ratio of $80:20$. We set the learning rate to $0.001$ and the batch size to $32$, and trained the model for $50$ epochs. We used Area Under Receiver Operating Curve (AUROC) and accuracy metrics to assess classification performance. Moreover, we calculated the $p$-values of our best-performing model against other models to show the statistical significance of the results. For a fair comparison, we used the same training settings and data partitioning for comparing methods~\cite{kusunose2023deep,schlesinger2022deep,li2023towards,wu2018multimodal}. An Nvidia RTX4090 GPU was used for all experiments. The implementation of all the models was carried out in Python (version $3.10$) with PyTorch~\cite{paszke2019pytorch}.

\noindent\textbf{Unimodal Study:} Table~\ref{my-reorganized-label} compared the results of unimodal models for CXR and ECG in rows $2-7$. We considered Kusunose et. al.'s method~\cite{kusunose2023deep} as a baseline for CXR modality and Schlesinger et al.'s method~\cite{schlesinger2022deep} as a baseline for ECG. Our $\text{CardioVAE}_\text{X,G}$ fine-tuning on unimodal data outperformed other unimodal methods for both modalities. The results show that $\text{CardioVAE}_\text{X,G}$ obtains improvements of $\Delta$AUROC $=0.100$ and $\Delta$Accuracy $=0.014$ over the CXR baseline~\cite{kusunose2023deep}, and the improvements of $\Delta$AUROC $=0.074$ and $\Delta$Accuracy$=0.024$ over ECG baseline~\cite{schlesinger2022deep}. The baseline methods do not use the unsupervised pre-training of models. Our $\text{CardioVAE}_\text{X,G}$ leverages pre-training from a large unlabeled dataset, and learns modality-specific features for the inference, enabling it to achieve better performance. To show the effect of pre-training, unimodal CNN models (row $1$ and $4$) are included in Table~\ref{my-reorganized-label} which use the same encoders and classification layers as in our $\text{CardioVAE}_\text{X,G}$ without pre-training. The results show that pre-training is important in achieving higher performance. We also included the current baseline model for high-cost cardiac MRI unimodal for PAWP prediction (row $7$), which was tested on a different cardiac MRI cohort and not for direct comparison with our low-cost unimodal models.

%We also designed a traditional machine learning-based method for the comparison, which utilizes Principal Component Analysis (PCA) and Linear Support Vector Machine (SVM), following prior works~\cite{tripathi2023tensor,swift2021machine}.

\definecolor{lightgray}{gray}{0.9}

\begin{table}[t]
\centering
\caption{Performance comparison using two metrics (with \textbf{best} in bold). CM: Cardiac Measurements from cardiac MRI, CMRI (4ch): Four-chamber cardiac MRI. Garg et al.~\cite{garg2022cardiac}\textsuperscript{\ding{169}} and Tripathi et al.~\cite{tripathi2023tensor}\textsuperscript{\ding{169}} were tested on a different cohort and included here for reference only. }
\label{my-reorganized-label}
\scalebox{0.84}{
\begin{tabular}{l|l|l|c|c|c}
\hline
 & Modality(s) & Method & AUROC & $p$-value$_{\text{AUC}}$ & Accuracy \\ \hline
\multirow{7}{*}{Unimodal} & \multirow{3}{*}{CXR} & CNN & $0.638 \pm 0.06$ & $< 0.0001$ & $0.675 \pm 0.03$ \\
& & Kusunose et al.~\cite{kusunose2023deep} & $0.581 \pm 0.04$ & $< 0.0001$ & $0.695 \pm 0.06$  \\
& & $\text{CardioVAE}_\text{X,G}$ (ours) & $\mathbf{0.681 \pm 0.05}$ & $< 0.0001$ & $\mathbf{0.709 \pm 0.04}$  \\
\cline{2-6}
& \multirow{3}{*}{ECG} & CNN & $0.724 \pm 0.05$ & $< 0.0021$ & $0.707 \pm 0.05$  \\
& & Schlesinger et al.~\cite{schlesinger2022deep} & $0.670 \pm 0.03$ & $< 0.0001$ & $0.703 \pm 0.04$ \\
& & $\text{CardioVAE}_\text{X,G}$ (ours) & $\mathbf{0.744 \pm 0.05}$ & $0.0226$ & $\mathbf{0.727 \pm 0.04}$ \\
\cline{2-6}
&\cellcolor{lightgray}CM & \cellcolor{lightgray}Garg et al.~\cite{garg2022cardiac}\textsuperscript{\ding{169}} & \cellcolor{lightgray}$0.730 \pm 0.04$ & \cellcolor{lightgray}- & \cellcolor{lightgray}$0.740 \pm 0.03$ \\
\hline
\multirow{5}{*}{Multimodal}
& \cellcolor{lightgray}CMRI (4ch) \& CM & \cellcolor{lightgray}Tripathi et al.~\cite{tripathi2023tensor}\textsuperscript{\ding{169}}  & \cellcolor{lightgray}$0.813 \pm 0.02$ &\cellcolor{lightgray} - &  $\cellcolor{lightgray}0.792 \pm 0.02$  \\
\cline{2-6}
& \multirow{5}{*}{CXR \& ECG} & CNN & $0.748 \pm 0.05$ & $0.0352$ & $0.735 \pm 0.03$  \\
& & Li et al.~\cite{li2023towards} & $ 0.737 \pm 0.05$ & $0.0101$ & $0.724\pm 0.04$ \\
& & Wu et al.~\cite{wu2018multimodal} & $0.758 \pm 0.03$ & $0.0283$ & $0.756 \pm 0.04$ \\
& & $\text{CardioVAE}_\text{X,G}$ (ours) & $\mathbf{0.790 \pm 0.03}$ & - & $\mathbf{0.772 \pm 0.04}$  \\
\hline
\end{tabular}}
\end{table}

\noindent\textbf{Multimodal Study}:
We compared $\text{CardioVAE}_\text{X,G}$ fine-tuned with multimodal (CXR \& ECG) data against four competing methods (rows $8-11$) in Table~\ref{my-reorganized-label}. Li~et~al.~\cite{li2023towards} used feature concatenation to combine two modalities in their multimodal variational autoencoder. Wu~et~al.~\cite{wu2018multimodal} utilized Product of Expert (PoE) based fusion. These two methods are based on a single-stream approach. Our $\text{CardioVAE}_\text{X,G}$ outperformed these two methods. Therefore, the tri-stream strategy in our model is effective for learning unique modality-specific along with shared features for prediction.  Additionally, obtained $p$-values show that our best-performing model produces statistically significant results against other models. We also included the comparison with the multimodal pipeline in Tripathi~et~al.~\cite{tripathi2023tensor} which was tested on a different cardiac MRI cohort. The results show that our model produces very competitive performance using low-cost data modalities. The scanning cost and time of cardiac MRI are very high in comparison to CXR and ECG modalities. Faster and easier scanning is vital for critical patients in clinical settings. Thus, PAWP prediction using CXR and ECG will be beneficial for clinicians. Next, our model also produces better results than multimodal CNN model (row $9$) which uses the same backbone as our model without multimodal pre-training, showing the potential of multimodal pre-training.

\begin{figure}[!t]
  \centering
  % Top row of subfigures for (a)
  \begin{subfigure}[b]{0.46\textwidth}
    \includegraphics[width=\textwidth]{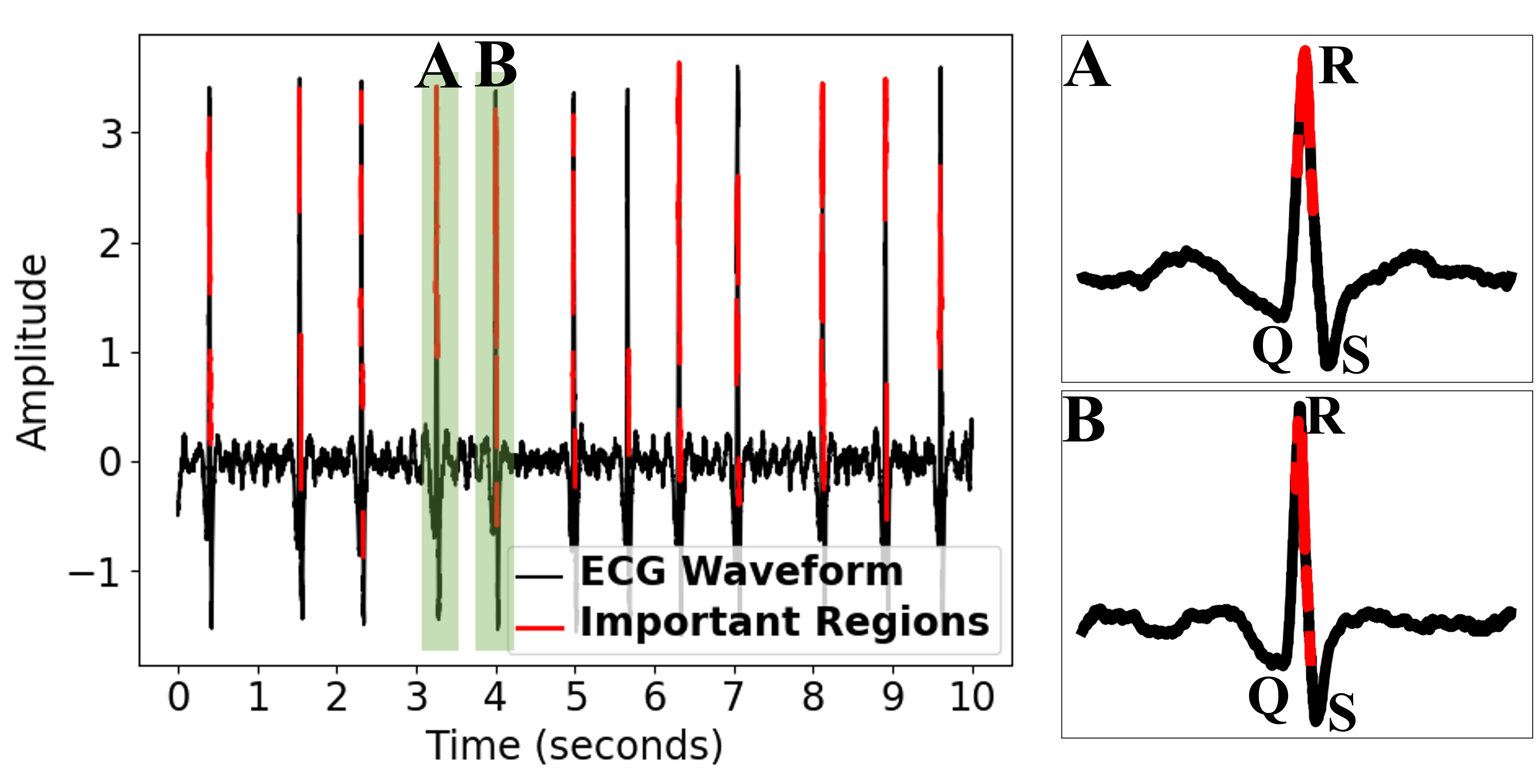}
    \label{fig:ecg_a}
  \end{subfigure}
  \hfill
  \begin{subfigure}[b]{0.46\textwidth}
    \includegraphics[width=\textwidth]{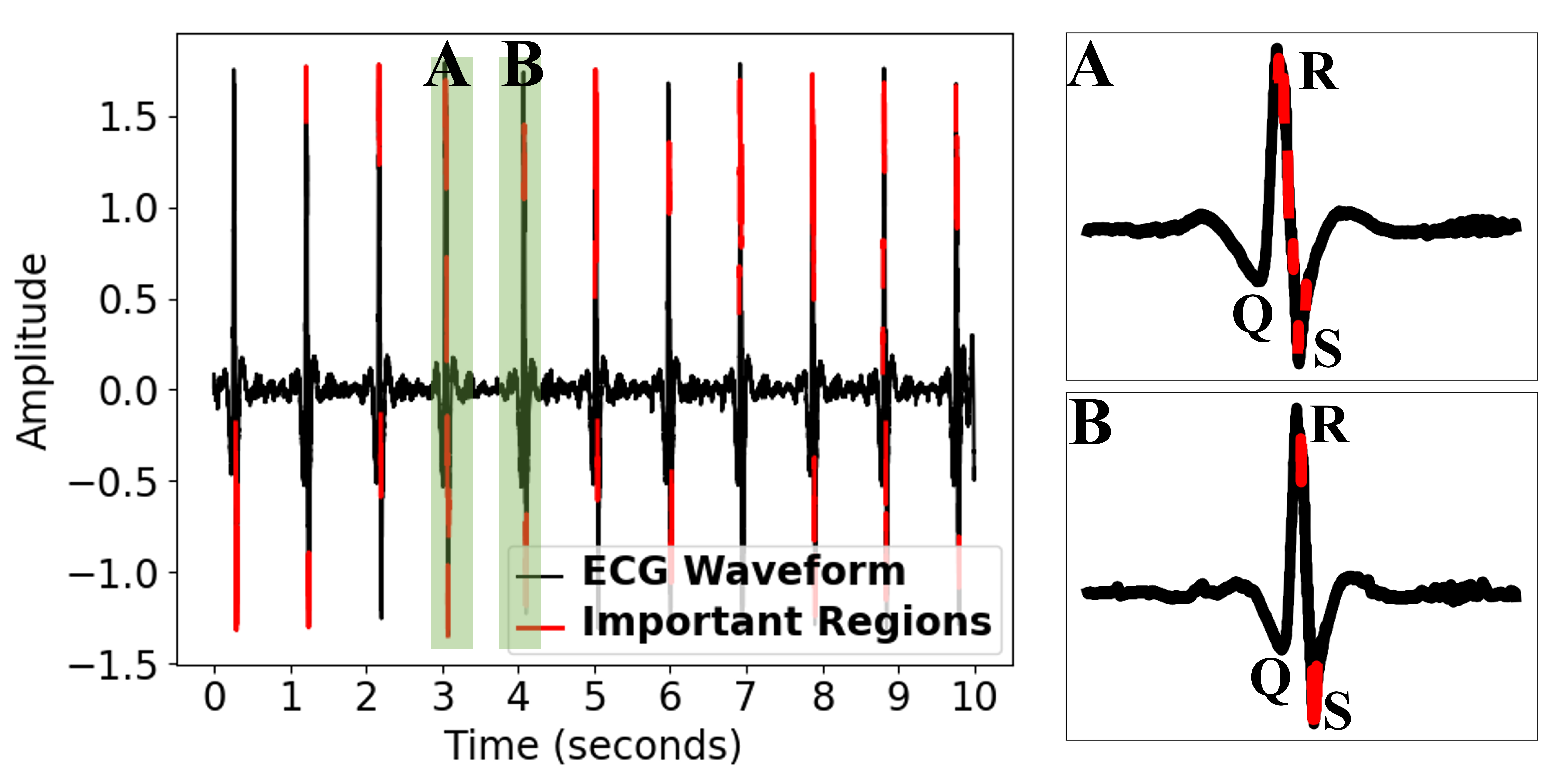}
    \label{fig:xray_a}
  \end{subfigure}
  
  % Bottom row of subfigures for (b)
  \hspace{0.04\textwidth}
  \begin{subfigure}[b]{0.40\textwidth}
    \includegraphics[width=\textwidth]{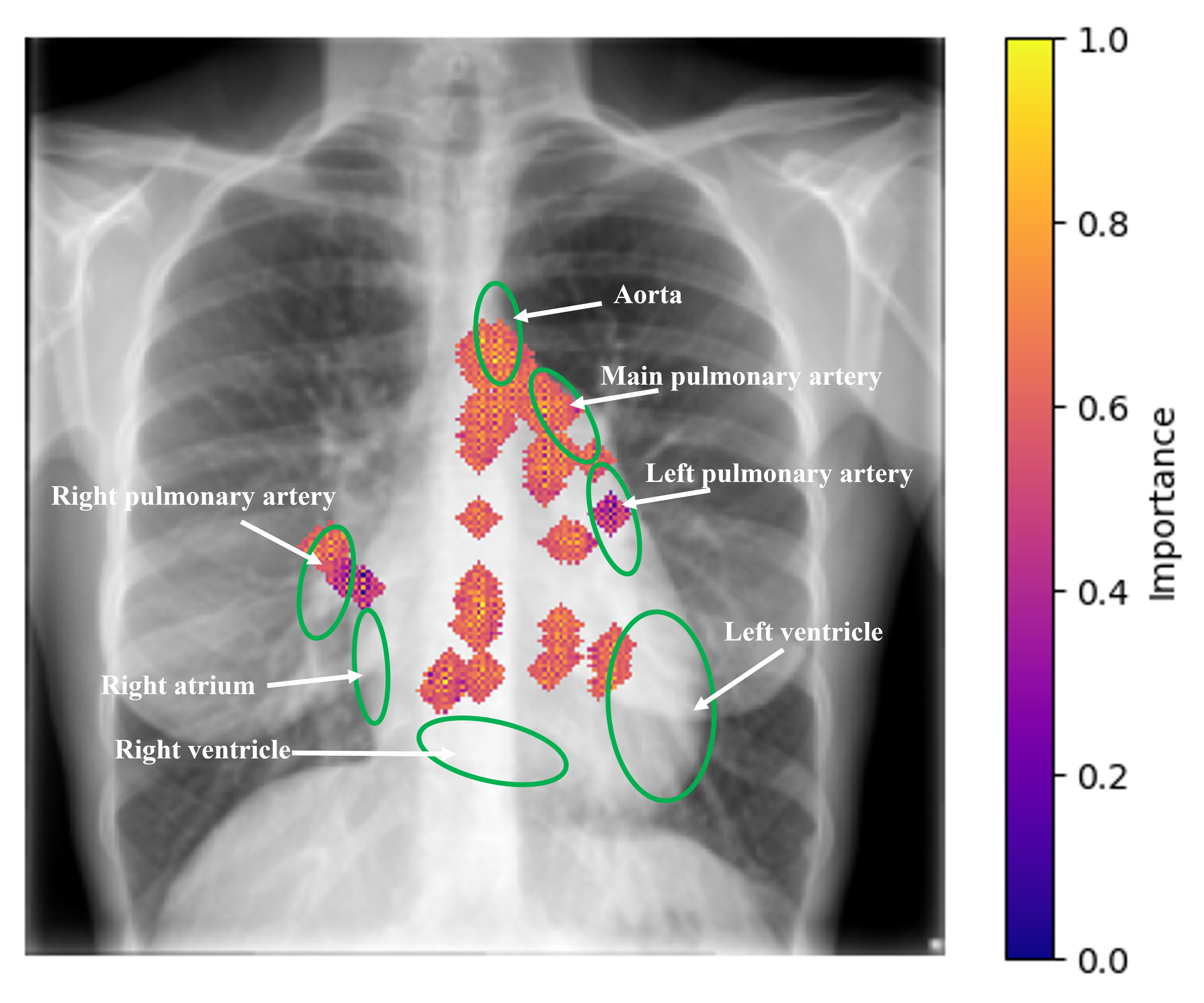}
    \caption{Normal PAWP Patient}
    \label{fig:ecg_b}
  \end{subfigure}
  \hfill
  \begin{subfigure}[b]{0.40\textwidth}
    \includegraphics[width=1\textwidth]{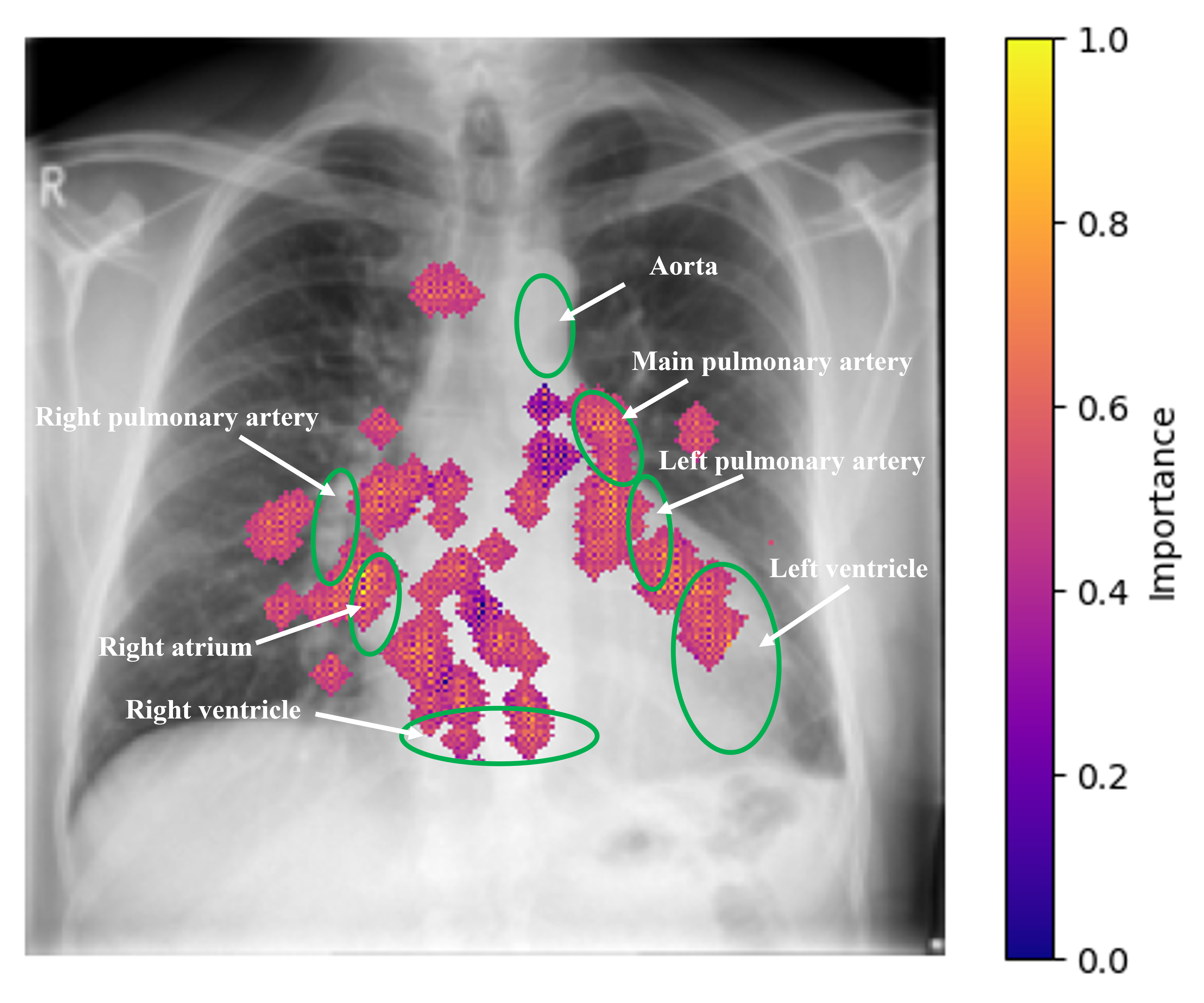}
    \caption{Elevated PAWP Patient}
    \label{fig:xray_b}
  \end{subfigure}
  
  \caption{Interpretability of $\text{CardioVAE}_\text{X,G}$ for two subjects using integrated gradients method~\cite{sundararajan2017axiomatic}. (a) 1D ECG (top left) and CXR (bottom left) for normal PAWP, (b) 1D ECG (top right) and CXR (bottom right) for elevated PAWP. Green annotations on CXRs highlight seven regions of the heart and lungs, marked by an expert clinician for enhanced visualization of key areas. The 1D ECG signal was smoothed with NeuroKit2~\cite{makowski2021neurokit2} library for better visualization.}
  \label{fig:both_interpret}
\end{figure}

%\noindent\textbf{Ablation Study:} We studied the effect of pre-training on unlabelled data. In Table~\ref{ablation}, we compared two models: a CNN and our $\text{CardioVAE}_\text{X,G}$. The CNN model uses the same encoders and FC layers (as in $\text{CardioVAE}_\text{X,G}$) but it does not use pre-training. The results show that pre-training improves performance for unimodal and multimodal predictions. Our best $\text{CardioVAE}_\text{X,G}$ model archives improvements of $\Delta$AUROC$=0.037$ and $\Delta$Accuracy$=0.014$ over the best CNN model.

%\begin{table}[!t]
%\centering
%\caption{Effect of pre-training with unlabeled data. CNN uses the same encoders %and classification layers as in our $\text{CardioVAE}_\text{X,G}$ but without pre-training.}
%\label{ablation}
%\begin{tabular}{l@{\hspace{6pt}}|@{\hspace{6pt}}l@{\hspace{6pt}}|@{\hspace{6pt}}c@{\hspace{6pt}}|@{\hspace{6pt}}c@{\hspace{6pt}}}
%\hline
%Method & Modality & AUROC & Accuracy \\ \hline
%\multirow{3}{*}{CNN} & CXR & $0.638 \pm 0.06$ & $0.675 \pm 0.03$ \\
 %& ECG & $0.724 \pm 0.05$ & $0.707 \pm 0.05$ \\
 %& CXR, ECG & $0.748 \pm 0.05$ & $0.735 \pm 0.03$ \\ \hline
%\multirow{3}{*}{$\text{CardioVAE}_\text{X,G}$ (ours)} & CXR & $0.681 \pm 0.05$ & $0.709 \pm 0.04$ \\
% & ECG & $0.744 \pm 0.05$ & $0.727 \pm 0.04$ \\
% & CXR, ECG & $0.790 \pm 0.03$ & $0.772 \pm 0.04$ \\ \hline
%\end{tabular}
%\end{table}

\noindent\textbf{Model Interpretation:} We used the integrated gradients method~\cite{sundararajan2017axiomatic} to demonstrate the interpretability of our best performing $\text{CardioVAE}_\text{X,G}$ model with both CXR and ECG modalities. Fig.~\ref{fig:both_interpret} depicts important regions for a normal subject and an abnormal subject, as identified by our model's decisions. For ECG, the model focuses on $R$ peak (for normal PAWP) as shown in zoomed-in segments, whereas the model relies on $R$ and $S$ peaks for abnormal PAWP. This indicates that our model performs the prediction based on QRS complex region~\cite{kashani2005significance}. The distinct alterations in the QRS complex enable the identification of left ventricular structural changes and conduction abnormalities, which are closely linked to variations in PAWP, reflecting the heart's response to altered cardiac hemodynamic states. In CXR images, the model focuses on cardiac regions, i.e. the left and right ventricles and arteries. By examining these regions in CXR, the model identifies their enlargement or structural changes, important indicators of cardiac function and fluid status. This offers important insights into PAWP levels by detecting subtle radiographic features of cardiac hemodynamic shifts and ventricular pressure alterations.

\section{Conclusion and Future Work}

This paper presented a multimodal variational autoencoder for CHDI detection from CXR and ECG. We showed that 1) the low-cost medical modalities (i.e., CXR and ECG) can be used to detect CHDI and are comparable to~\cite{garg2022cardiac,tripathi2023tensor} from high-cost modalities such as cardiac MRI, 2) the employed tri-stream unsupervised pre-training improved the performance of unimodal and multimodal models compared to~\cite{kusunose2023deep,schlesinger2022deep,li2023towards,wu2018multimodal}, and 3) interpretations made by our model are relevant for clinical decision-making as confirmed by a clinician. Future work can extend $\text{CardioVAE}_\text{X,G}$ to other cardiac hemodynamics prediction tasks. 

%our model can produce interpretable predictions for CXR and ECG modalities. In the future, we would like to explore the applicability of our model for other cardiac hemodynamics prediction tasks.

%Future work will focus on expanding our models to include additional data modalities, such as textual data and Electronic Health Records (EHR), as well as more downstream tasks like mortality and Pulmonary Hypertension prediction. This will allow us to explore the applicability for our CardioVAE method across various modalities and tasks. 

%\section*{Acknowledgment}

%
\bibliographystyle{splncs04}
\bibliography{mybibliography}

\end{document}